\documentclass[10pt,twocolumn,letterpaper]{article}
\usepackage{cvpr}              
\usepackage[dvipsnames]{xcolor}

\definecolor{cvprblue}{rgb}{0.21,0.49,0.74}
\usepackage[pagebackref,breaklinks,colorlinks,citecolor=cvprblue]{hyperref}
\usepackage{times}
\usepackage{epsfig}
\usepackage{graphicx}
\usepackage{amsmath}
\usepackage{amssymb}
\usepackage{bm,multirow,arydshln,subcaption,booktabs,makecell}

\def\paperID{9064} 
\def\confName{CVPR}
\def\confYear{2024}

\title{Meta-Point Learning and Refining for Category-Agnostic Pose Estimation}

\author{Junjie Chen\textsuperscript{\rm 1}~~~~~~Jiebin Yan\textsuperscript{\rm 1}\thanks{Corresponding author}~~~~~~Yuming Fang\textsuperscript{\rm 1}~~~~~~Li Niu\textsuperscript{\rm 2}\\
\textsuperscript{\rm 1} Jiangxi University of Finance and Economics, \\
\textsuperscript{\rm 2} Shanghai Jiao Tong University\\
{\tt\small \{chenjunjie,yanjiebin,fangyuming\}@jxufe.edu.cn,} \tt\small {ustcnewly@sjtu.edu.cn}
}

\begin{document}
\maketitle
\begin{abstract}
Category-agnostic pose estimation (CAPE) aims to predict keypoints for arbitrary classes given a few support images annotated with keypoints.
Existing methods only rely on the features extracted at support keypoints to predict or refine the keypoints on query image, but a few support feature vectors are local and inadequate for CAPE.
Considering that human can quickly perceive potential keypoints of arbitrary objects, we propose a novel framework for CAPE based on such potential keypoints (named as meta-points).
Specifically, we maintain learnable embeddings to capture inherent information of various keypoints, which interact with image feature maps to produce meta-points without any support.
The produced meta-points could serve as meaningful potential keypoints for CAPE.
Due to the inevitable gap between inherency and annotation, we finally utilize the identities and details offered by support keypoints to assign and refine meta-points to desired keypoints in query image.
In addition, we propose a progressive deformable point decoder and a slacked regression loss for better prediction and supervision.
Our novel framework not only reveals the inherency of keypoints but also outperforms existing methods of CAPE.  
Comprehensive experiments and in-depth studies on large-scale MP-100 dataset demonstrate the effectiveness of our framework.
Code is avaiable at \url{https://github.com/chenbys/MetaPoint}
\end{abstract}
\section{Introduction} \label{sec:intro}
Pose estimation is a fundamental and significant computer vision task, which aims to produce the locations of pre-defined semantic part of object instance in 2D image.
Recently, it has received increasing attention in the computer vision community due to its wide applications in virtual reality, augmented reality, human-computer interaction, robot and automation.
However, most pose estimation methods are trained with category-specific data and thus cannot be applied for novel classes, especially when they have different keypoint classes.
Therefore, category-agnostic pose estimation (CAPE) \cite{POMNet,CapeFormer} is recently proposed to localize desired keypoints for arbitrary classes given one or few support images annotated with keypoints.

\begin{figure}[t]
\begin{center}
\includegraphics[width=1\linewidth]{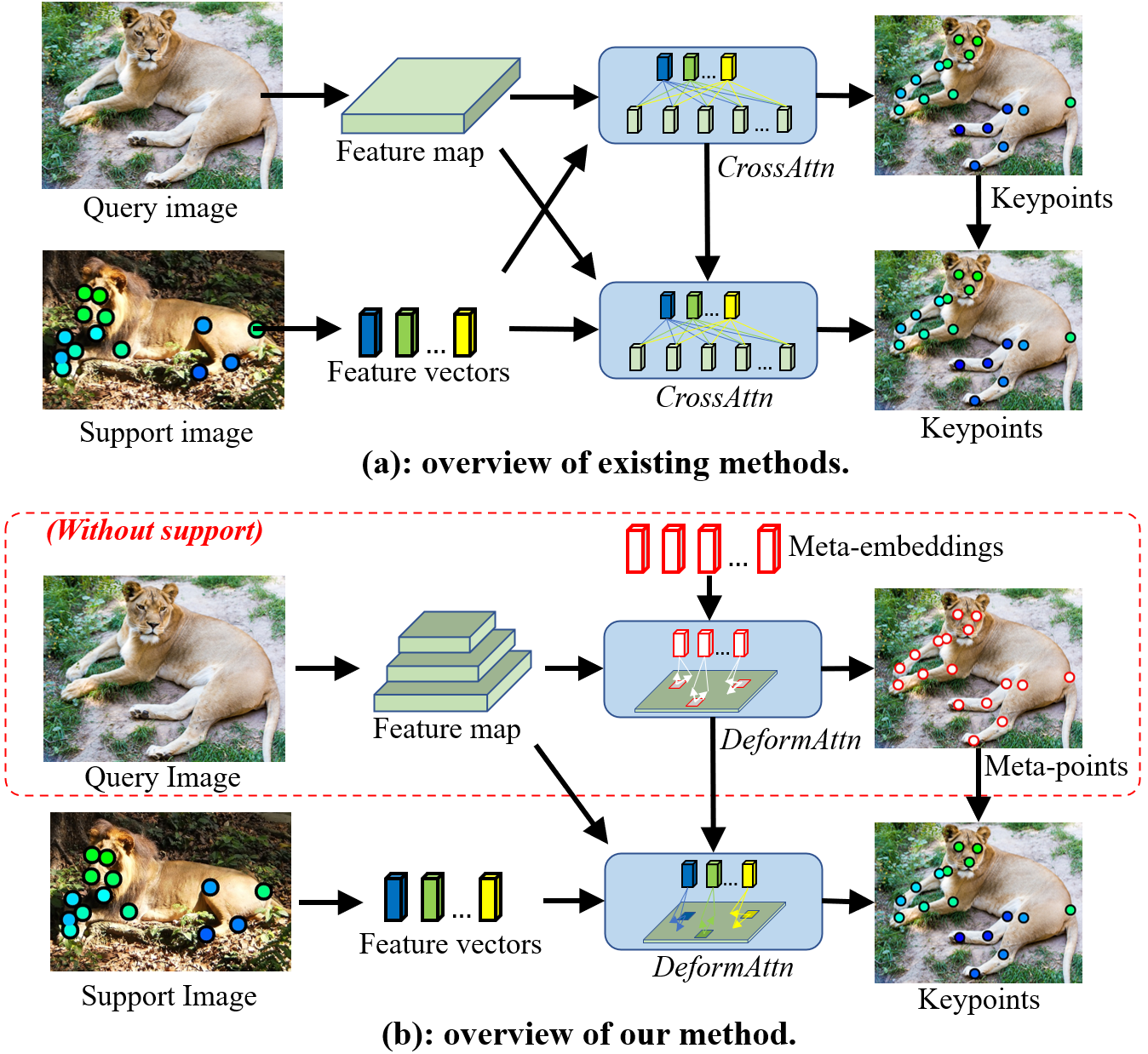}
\end{center}
\caption{Overview of existing methods and our method.
(a): Existing methods only rely on the local features at support keypoints to predict or refine the keypoints.
(b): Our method employs learnable embeddings to capture inherent information and produces meta-points without support. 
Then our method assigns and refines meta-points according to support keypoints.
}
\label{fig:overview}
\end{figure}

Unlike the standard pose estimation task where the keypoint classes are pre-defined, the desired keypoints in CAPE are defined by support keypoints.
Therefore, existing methods \cite{POMNet,CapeFormer} intuitively extract features at support keypoints and then rely on them to predict or refine the keypoints on query image, as shown in Fig.~\ref{fig:overview}~(a).
Although such straightforward pipeline has achieved considerable success, the information provided by a few feature vectors are local \cite{nonlocal,nonlocal2} and inadequate, especially when support keypoints are partially occluded or blurred.
Furthermore, above methods focus on learning pixel-level correspondences and neglect to mine inherent and complementary information of keypoints, which is a fundamental issue of CAPE.

We conjecture that there are some inherent or universal points in each object, based on the insight that human can quickly grasp the essentials of keypoints of arbitrary objects without any reference or support image.
We refer to such class-agnostic potential points as meta-points, which may seem similar to object proposal \cite{PropBox1,PropBox2,PropBox3} or mask proposal \cite{PropMaskDeep,PropMask1,PropMask2} but actually have significant characteristics.
For single object, meta-points could provide fine-grained and structural information with a sparse representation.
For two objects, meta-points of the same semantic part should have consistent contact.
For example, a lion and a bird may have consistent meta-points (\emph{e.g.}, eye) and individual meta-points (\emph{e.g.}, wing). 
Therefore, meta-point learning is a fundamental yet challenging issue.

In this paper, we propose a novel pipeline for CAPE, which first predicts meta-points without support and then assign and refine them to desired keypoints according to support, as illustrated in Fig.~\ref{fig:overview}~(b).
Specifically, we maintain learnable meta-embeddings to store the inherent information of various keypoints for producing meta-points.
Given a query image, these meta-embeddings will interact with its feature maps via a transformer decoder to mine the inherent information and thus could produce meta-points without support.
To bridge the gap between inherent meta-points and desired keypoints, we first employ bipartite matching-based assignment to assign optimal meta-points for desired keypoints according to support.
Afterwards, we employ the support feature vectors together with the mined inherent information to refine assigned meta-points via another transformer decoder.
Thanks to the meta-points and mined inherent information, our framework could produce more precise keypoints for CAPE.
In addition, we propose a progressive deformable point decoder, which progressively mines fine-grained features to decode points with last points as reference based on deformable attention \cite{DeformDETR}.
We also propose a slacked regression loss to reduce the immature gradients of auxiliary decoder layers.

We evaluate our framework on the largest dataset for CAPE, \emph{i.e.}, Multi-category Pose (MP-100) dataset \cite{POMNet}.
The in-depth studies demonstrate that our method could learn meaningful meta-points for arbitrary classes without support.
And the comprehensive experiments indicate that our meta-points could dramatically facilitate CAPE and lead to better performance against existing methods.
Our contributions could be summarized as: 
\textbf{1)} To the best of our knowledge, we are the first to learn class-agnostic potential keypoints for CAPE. 
\textbf{2)} We propose a novel framework for CAPE based on meta-point learning and refining. 
\textbf{3)} Two minor contributions are progressive deformable point decoder and slacked regression loss for better prediction and supervision.
\textbf{4)} Our method not only produces meaningful meta-points but also predicts more precise keypoints against existing methods in CAPE.

\section{Related Works}
\subsection{Category-Specific Pose Estimation}
Pose estimation is a fundamental task in computer vision, which aims to localize the pre-defined keypoints of objects in the input image.
Most pose estimation methods are class-specific, \emph{e.g.}, estimating pose for human \cite{humanpose1,humanpose2}, animals \cite{animalpose1,animalpose2}, and vehicles \cite{carpose1,carpose2}.
Technically, existing pose estimation methods could be roughly categorized into regression-based methods \cite{reg1,reg_RLE,reg_DEKR}, heatmap-based methods \cite{heatmap1,heatmap2,heatmap3}, and query-based methods \cite{query_PETR,query1,query2}.
To name a few, Li \emph{et al.} \cite{reg_RLE} proposed a novel regression paradigm with Residual Log-likelihood Estimation to learn the change of the distribution to facilitate the training process.
Geng \emph{et al.} \cite{reg_DEKR} proposed to employ multi-branch structure for disentangled keypoint regression, which could be able to attend to the keypoint regions and thus improve the performances.
Luo \emph{et al.} \cite{heatmap_SAHR} proposed to estimate scale-adaptive heatmaps, which adaptively adjusts the standard deviation for each keypoint and thus is tolerant of various scales and ambiguities.
Shi \emph{et al.} \cite{query_PETR} proposed to learn multiple pose queries for reasoning a set of full-body poses and then use a joint decoder to refine by exploring kinematic relations.
Although above works have achieved great success for estimating pose of specific classes, they are inapplicable for novel classes, especially when the novel classes have different number and kinds of keypoints.
In this work, we focus on estimating pose for arbitrary classes with a few support images annotated with desired keypoints.

\subsection{Category-Agnostic Pose Estimation}
Few-shot learning \cite{fsl1} focuses on learning novel classes using only a few samples and has been applied in extensive visual tasks, \emph{e.g.}, classification \cite{fsl-cls1,m2}, object detection \cite{fsl-det1,fsl-det2}, segmentation \cite{fsl-seg1,fsl-seg2,m4}.
For keypoint estimation, previous few-shot methods mostly focus on specific domains, \emph{e.g.}, facial images \cite{face1,face2}, clothing images \cite{Metacloth}, or X-ray images \cite{x0,x1}.
For wider classes, Lu \emph{et al.} \cite{FSKD} explored a flexible few-shot scenario including novel/base classes and novel/base keypoints.
Later, Lu \emph{et al.} \cite{SFSKD} proposed a saliency-guided transformer for above scenario and transductive scenario.
He \emph{et al.} \cite{FewShot3DKP} explored to estimate keypoints using a few examples combined with unlabeled images.
Recently, POMNet \cite{POMNet} elaborated a large-scale dataset including 100 classes and set the task of category-agnostic pose estimation.
Technically, POMNet \cite{POMNet} proposed a keypoint matching framework, which extracts the features at support keypoints and employs their matching similarities to every pixels to retrieve the desired keypoints.
Afterwards, CapeFormer \cite{CapeFormer} enhanced the similarity modeling in above matching pipeline, and proposed to further refine each keypoint via a elaborate transformer decoder.
Although above methods \cite{POMNet,CapeFormer} have great promoted the development of CAPE, they only rely on the local features at support keypoints and neglect to mine inherent and universal information of keypoints.
In contrast, we propose to mine the inherent or universal points in objects without any support, and propose a progressive deformable point decoder to progressively mines fine-grained features to decode points more precisely.

\subsection{Category-Agnostic Proposal Learning}
Category-agnostic proposal learning is a long-standing research topic in computer vision and has extensive downstream tasks.
Early methods rely on low-level features (\emph{e.g.}, edges, super-pixels, or saliency) to produce bounding box proposals \cite{PropBox1,PropBox2} or mask proposals \cite{PropMask1,PropMask2,m1,m3}.
Recent methods employ various network to learn box proposals \cite{PropBoxD1,FasterRCNN} or mask proposals \cite{PropMaskD1,PropMaskD2}.
To name a few, one of the most impressive works is Faster R-CNN \cite{FasterRCNN}, which first learns class-agnostic box proposals and then refines them to object bounding boxes.
Pinheiro \emph{et al.} \cite{PropMaskDeep} proposed to learn the class-agnostic segmentation masks and their object likelihood scores. 
Therefore, proposal learning is a significant and foundational task and could greatly facilitate downstream tasks in the second stage.
Nevertheless, there are few works on learning class-agnostic keypoints, and we propose to learn potential keypoints and name our learned points as meta-points. 
Basically, our meta-point could provide fine-grained and structural information with a sparse representation.
Furthermore, our meta-points have consistent contact for the same semantic part of different objects.
Therefore, such meta-points not only reveals the inherency of keypoints but also serves as meaningful proposals for downstream tasks.
Although the coarse keypoints in \cite{CapeFormer} are named as ``proposals'', their ``proposals'' are predicted according to support keypoints and thus do not have the class-agnostic properties.

\section{Method}
\begin{figure*}[h]
\begin{center}
\includegraphics[width=1\linewidth]{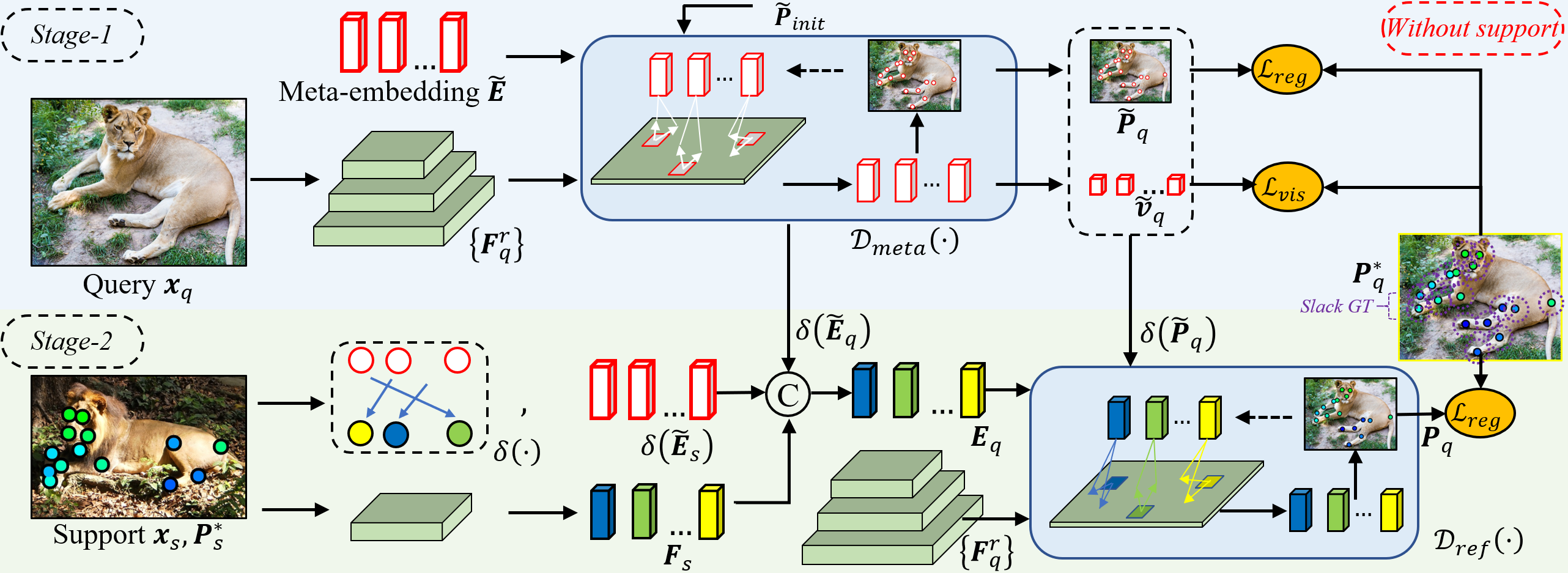}
\end{center}
\caption{Our framework employs two stages to predict meta-points and desired keypoints.
In the first stage, the learnable meta-embeddings interact with query feature maps via our progressive deformable point decoder to mine inherent information to predict meta-points and their visibilities.
In the second stage, the meta-points are assigned with identities according to the given support keypoints.
After that, the assigned meta-points are refined to desired keypoints based the support features and mined inherent information via another point decoder.
}
\label{fig:framework}
\end{figure*}

Class-agnostic pose estimation (CAPE) aims to estimate desired keypoints on query image given a few support images annotated with keypoints for novel classes.
For $N$-shot setting, CAPE could be formulated as a function:
\begin{equation}
    \bm{P}_q=\mathcal{F}_{cape}(\bm{x}_q,\{\bm{x}_{s_{n}}\}_{n=1}^N,\{\bm{P}^*_{s_{n}}\}_{n=1}^N)
\end{equation}
where $N$ support images $\{\bm{x}_{s_{n}}\}_{n=1}^N$ and their ground-truth (GT) keypoints $\{\bm{P}^*_{s_{n}}\}_{n=1}^N$ are given, and the task is to estimate the desired keypoints $\bm{P}_q$ on query image $\bm{x}_q$.
To evaluate the generalization capacity for arbitrary classes, all classes in benchmark are split into non-overlapping base classes and novel classes, and query and support images are sampled from only base (\emph{resp.}, novel) classes for training (\emph{resp.}, evaluating).
For simplicity, we first introduce our method in 1-shot setting, and then clarify the intuitive extension with more support images.
Specifically, the function is simplified as $\bm{P}_q=\mathcal{F}_{cape}(\bm{x}_q,\bm{x}_s,\bm{P}^*_s)$, where $\bm{x}_q,\bm{x}_s \in \mathbb{R}^{3\times H\times W}$ and $\bm{P}_q,\bm{P}^*_s \in \mathbb{R}^{K\times 2}$.
For brevity of description, we employ subscript to indicate the variable source and use square bracket to show the index in variable, \emph{e.g.}, $\bm{P}_q[k]$ for the $k$-th keypoint on query image.
To tackle such problem, we propose a novel framework based on meta-point learning and refining, which not only reveals the keypoint inherency but also produces precise keypoints.
The detailed framework is illustrated in Fig. \ref{fig:framework}, we first introduce the pipelines of two-stages and then clarify the architecture of our point decoder used in pipelines.

\subsection{Predicting Meta-Point without Support} \label{sec:pred_meta}
Considering that human is able to quickly abstract keypoints for arbitrary classes without support, we propose to model such fundamental ability to facilitate CAPE.
As shown in the upper part of Fig. \ref{fig:framework}, we maintains $M$ trainable meta-embeddings $\Tilde{\bm{E}} \in \mathbb{R}^{M\times D}$ in our model (\emph{e.g.}, $M=100$), which are shared by all samples and store the inherent and universal information of keypoints.
For the input sample $\bm{x}_q$, our model first employs ResNet-50 \cite{ResNet} backbone to extract pyramid feature maps $\{\bm{F}^r_q\}_{r=1}^{R}$, where $\bm{F}^r_q \in \mathbb{R}^{D^r\times H^r\times W^r}$ corresponds to the $r$-th residual layer in deep-to-shallow order ($R=3$ by default).
Afterwards, the meta-embeddings interact with image feature maps to mine inherent information, which could be summarized as:
\begin{equation}
\left \{\left[\Tilde{\bm{P}}^l_q; \Tilde{\bm{E}}^l_q\right]\right\}_{l=1}^{L}=\mathcal{D}^L_{meta}(\Tilde{\bm{E}}, \Tilde{\bm{P}}_{init}, \{\bm{F}^r_q\}_{r=1}^{R}).
\end{equation}
Specifically, $\mathcal{D}_{meta}$ is implemented with our Progressive Deformable Point Decoder, which employs DeformAttn\cite{DeformDETR} to progressively decode $L$ layers of points with last points as reference points and will be clearly introduced in Sec. \ref{sec:decoder}.
$\Tilde{\bm{P}}_{init} \in \mathbb{R}^{M\times 2}$ are the initial points, and we adopt uniform grid points to launch the decoding for meta-points.
From the $l$-th decoder layer in $\mathcal{D}_{meta}$, we obtain the query meta-points $\Tilde{\bm{P}}^l_q \in \mathbb{R}^{M\times 2}$ and query meta-embeddings $\Tilde{\bm{E}}^l_q \in \mathbb{R}^{M\times D}$. 
We use the predictions from the last layer as the proposals for following processes.
Because not all meta-points are visible for arbitrary objects, we employ a lightweight binary classifier to predict the visibilities $\Tilde{\bm{v}}_q \in \mathbb{R}^{M}$ according to the last query meta-embeddings.

In this way, we are able to produce meta-points $\Tilde{\bm{P}}^L_q$ and their visibilities $\Tilde{\bm{v}}_q$ without any support, which reveal the inherency of keypoints and serve as meaningful proposals to facilitate downstream tasks.

\subsection{Refining Meta-Point with Support}
Due to the inevitable gap between keypoint inherency and annotation, we have to refine above meta-points to desired keypoints $\bm{P}_q$ with GT support keypoints $\bm{P}^*_s$.
The support information includes two aspects, \emph{i.e.}, desired identity and desired detail, which will be described as follows.

\subsubsection{Identity Assignment}
Each annotated keypoint has its identity/index, and thus we need to determine the index mapping $\delta(\cdot)$ between meta-points and desired keypoints, \emph{i.e.}, the $\delta(k)$ meta-point corresponds to the $k$-th keypoint.
Specifically, we employ the bipartite matching-based assignment as in \cite{MaskFormer,DETR} to solve the mapping according to a cost matrix $\mathcal{C}\in \mathbb{R}^{M\times K}$, where each entry $\mathcal{C}[m,k]$ indicates the cost for assigning the $m$-th meta-point to the $k$-th annotated keypoint.
To solve the assignment according to support, we obtain the meta-points $\{\Tilde{\bm{P}}^l_s\}_{l=1}^L$ and visibilities $\Tilde{\bm{v}}_s$ on support image as the same pipeline in Sec. \ref{sec:pred_meta}.
Then, we compute the cost by:
\begin{equation}\label{eqn:cost}
    \mathcal{C}_s[m,k]= \sum_{l=1}^L{\mathcal{L}_{sl_1}(\Tilde{\bm{P}}^l_s[m],\bm{P}^*_s[k],l) -\alpha {\rm log}(\Tilde{\bm{v}}_s[m])},
\end{equation}
where $\mathcal{L}_{sl_1}$ measures the $l_1$ deviation with slacks on auxiliary decoder layers and will be introduced in Sec. \ref{sec:loss}.
The second term considers the classification of meta-point visibility, with a trade-off hyper-parameter $\alpha=0.5$.
Finally, we use $\mathcal{C}_s$ to obtain the optimal assignment $\delta(\cdot)$.
In this way, each desired keypoint is assigned with a meta-point, \emph{i.e.}, the $k$-th keypoint is assigned with $\Tilde{\bm{P}} [\delta(k)]$.
Because we set $K \leq M$, and the unassigned meta-points are treated as ignore.
For brevity, we employ $\delta(\Tilde{\bm{P}})\in \mathbb{R}^{K\times 2}$ and $\delta(\Tilde{\bm{E}}) \in \mathbb{R}^{K\times D}$ to denote the assigned meta-points and meta-embeddings re-ordered in desired sequence.

\subsubsection{Details Enhancement}
Besides the desired identities, we refine meta-points to desired coordinates according to the support information.
By default, we employ the prediction of the last decoder layer in $\mathcal{D}_{meta}$ to refine and omit the decoder layer index for brevity, \emph{i.e.}, we use assigned meta-points $\delta(\Tilde{\bm{P}}_q)$ and their embeddings $\delta(\Tilde{\bm{E}}_q)$ to launch the refinement.
Basically, we derive support keypoint features $\bm{F}_s$ via ``soft'' ROI pooling as in previous works \cite{CapeFormer,POMNet}, which contain information of local details in support image.
We also collect the meta-embeddings assigned to support keypoints $\delta(\Tilde{\bm{E}}_s)$, which have absorbed the global inherent information in support image.
Afterwards, we squeeze the concatenation of above embeddings to obtain comprehensive embeddings $\bm{E}_q \in \mathbb{R}^{K\times D}$ for keypoints, \emph{i.e.}, $\bm{E}_q =f_{sqz}([\delta(\Tilde{\bm{E}}_q);\delta(\Tilde{\bm{E}}_s);\bm{F}_s])$.
Then, the assigned meta-points $\delta(\Tilde{\bm{P}}_q)$ are progressively refined according to $\bm{E}_q$ and image feature maps, as:
\begin{equation}
\left \{\left[\bm{P}^l_q; \bm{E}^l_q\right]\right\}_{l=1}^{L}=\mathcal{D}^L_{ref}(\bm{E}_q, \delta(\Tilde{\bm{P}}_q), \{\bm{F}^r_q\}_{r=1}^{R}).
\end{equation}
Specifically, we also employ our Progressive Deformable Point Decoder to implement $\mathcal{D}_{ref}$ (same architecture but different parameters with $\mathcal{D}_{meta}$) and leave the details to Sec. \ref{sec:decoder}.
In this way, our decoder $\mathcal{D}_{ref}$ progressively refines assigned meta-points to desired keypoints.

\subsection{Progressive Deformable Point Decoder} \label{sec:decoder}
\begin{figure}[h]
\begin{center}
\includegraphics[width=0.95\linewidth]{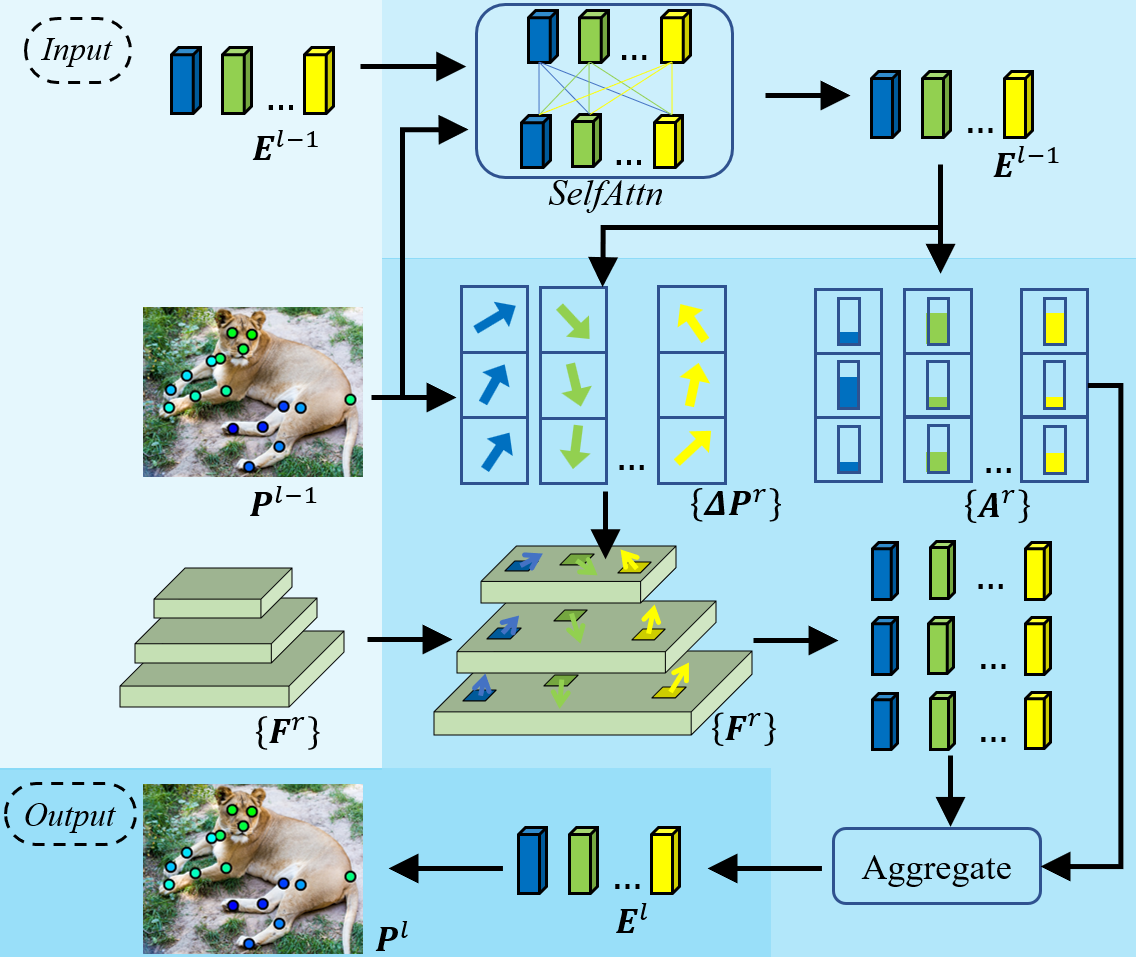}
\end{center}
\caption{
Illustration of decoder layer in our Progressive Deformable Point Decoder.
The input embeddings first interact with each other via a self-attention module.
After that, the offsets and weights are predicted to mine fine-grained features on input feature maps.
Finally, a deformable attention module uses input points as reference to refine the embeddings and points.
}
\label{fig:decoder}
\end{figure}
Predicting precise keypoints requires fine-grained and high-resolution features, and thus we propose a progressive deformable point decoder, which makes full use of deformable attention module \cite{DeformDETR} to precisely refine keypoints.
Our decoder consists of $L$ decoder layers ($L=3$ by default), and the architecture of decoder layer is shown in Fig. \ref{fig:decoder}.
Given points $\bm{P}^{l-1}\in \mathbb{R}^{K\times 2}$ and their embeddings $\bm{E}^{l-1}\in \mathbb{R}^{K\times D}$, each decoder layer performs self attention \cite{multiheadattention} and deformable attention \cite{DeformDETR} to update points and embeddings according to feature maps $\{\bm{F}^r\}_{r=1}^R$, as: 
\begin{equation}
    [\bm{P}^{l};\bm{E}^{l}]=\mathcal{D}(\bm{E}^{l-1},\bm{P}^{l-1},\{\bm{F}^r\}_{r=1}^R).
\end{equation}
By stacking multiple such layers, our decoder could update the points in a progressive and precise manner.
Specifically, each decoder layer first employs a self attention module to render each point aware of other points' positions and contents as in \cite{CapeFormer} (also with identity embeddings and position embeddings).
Afterwards, each embedding predicts the offsets $\{\Delta \bm{ P}^r\}_{r=1}^R$ and weights $\{\bm{A}^r\}_{r=1}^R$ for paying attention to each feature map.
By using the points $\bm{P}^{l-1}$ as reference points, the deformable attention module aggregates sampled features according to weights and obtain updated embeddings $\bm{E}^l$.
Then, the coordinates of points are refined by: 
\begin{equation}
    \bm{P}^l=\sigma \left(\sigma^{-1}(\bm{P}^{l-1})+\mathcal{F}_{mlp}(\bm{E}^l) \right),
\end{equation}
where $\mathcal{F}_{mlp}$ is a lightweight point head and $\sigma$ indicates Sigmoid function. 
In this way, input points and their embeddings could be progressively refined according to the nearby fine-grained features. 

\subsection{Training and Inference} \label{sec:loss}
Our full training objective consists of keypoint regression loss and visibility classification loss.
For the regression loss, strict supervision on auxiliary decoder layers will trigger immature gradients for fitting keypoints with inadequate capacities.
And thus we adopt a slacked $L_1$ loss: 
\begin{equation}
    \mathcal{L}_{sl_1}(\bm{P}, \bm{P}^*,l)=\sum_k \varepsilon \left( \left|{\bm{P}[k]-\bm{P}^*[k]} \right|_1  -\lambda \cdot (L-l) \right), 
\end{equation}
where $\varepsilon$ indicates ReLU function and $\lambda=0.1$ is a slack hyper-parameter.
And our total regression loss is 
\begin{equation}
    \mathcal{L}_{reg}= \sum_{l=1}^{L}{\mathcal{L}_{sl_1}(\delta(\Tilde{\bm{P}}^l_q), \bm{P}^*_q,l)+\mathcal{L}_{sl_1}(\bm{P}^l_q, \bm{P}^*_q,l)},
\end{equation}
where the first part supervises the assigned meta-points and the second part supervises the refined meta-points towards to GT keypoints $\bm{P}^*_q$.
For the visibility classification loss, we employ binary cross-entropy loss:
\begin{equation}
    \mathcal{L}_{vis}=BCE(\Tilde{\bm{v}_q}, \bm{1}_{\delta}),
\end{equation}
where $\bm{1}_{\delta}\in \mathbb{R}^{M}$ is a vector indicating assigned meta-points with $1$, otherwise $0$.
Therefore, our full training objective is calculated as $\mathcal{L}_{full}=\mathcal{L}_{reg}+\alpha\cdot \mathcal{L}_{vis}$, where we reuse the balance hyper-parameter $\alpha$ in cost computation Eqn. \ref{eqn:cost}.

In the inference stage, we employ the result of last decoder layer $\bm{P}^L_q$ as the estimated keypoints.
As for $N$-shot setting, we average the support keypoint features from different shots as previous works \cite{CapeFormer,POMNet}.
Besides, we average their cost matrices to obtain better assignment $\delta(\cdot)$, \emph{i.e.}, $\mathcal{C}_{\Bar{s}}=\frac{1}{N}\sum_n^N{ \mathcal{C}_{s_{n}}}$.
We also average their assigned meta-embeddings, \emph{i.e.}, $\delta(\Tilde{\bm{E}}_{\Bar{s}})=\frac{1}{N}\sum_n^N{ \delta(\Tilde{\bm{E}}_{s_{n}}})$.
In this way, we could obtain enhanced information to facilitate prediction.
\section{Experiments}
\subsection{Dataset, Metric, and Implementation Details}
Overall, we follow the experimental setting of previous work \cite{CapeFormer}.
Specifically, we conduct experiments on MP-100 dataset \cite{POMNet}, which covers $100$ classes and $8$ super-classes and is the largest benchmark dataset for CAPE.
MP-100 contains over $18$K images and $20$K annotations, and the keypoint numbers range from $8$ to $68$ across different classes.
The collected $100$ classes are split into non-overlapping train/val/test sets with the ratio of $70:10:20$.
There are five random splits to reduce the impact of randomness.
We use the Probability of Correct Keypoint (PCK) \cite{PCK} as the quantitative metric.
Apart from the PCK of threshold $0.2$ as in \cite{POMNet,CapeFormer}, we also report the mPCK (of $[0.05,0.1,0.15,0.2]$) for a more comprehensive evaluation.

\begin{table*}[t]
\centering
\caption{Results of 1-shot and 5-shot setting on MP-100 dataset.
We summarize PCK@0.2 results and report mPCK results in Tab.~ \ref{tab:mPCK}.}
\label{tab:SOTA}
\scalebox{0.9}{
\begin{tabular}{c|cccccc|cccccc} \toprule
\multirow{2}{*}{Method} & \multicolumn{6}{c|}{1-shot setting} & \multicolumn{6}{c}{5-shot setting} \\
 & Split1 & Split2 & Split3 & Split4 & Split5 & Average &Split1 & Split2 & Split3 & Split4 & Split5 & Average \\ \midrule
ProtoNet \cite{ProtoNet}&46.05&40.84&49.13&43.34&44.54&44.78&60.31&53.51&61.92&58.44&58.61&58.56\\
MAML \cite{MAML} &68.14&54.72&64.19&63.24&57.20&61.50&70.03&55.98&63.21&64.79&58.47&62.50\\
Fine-tune \cite{Finetuning}&70.60&57.04&66.06&65.00&59.20&63.58&71.67&57.84&66.76&66.53&60.24&64.61\\
POMNet \cite{POMNet}&84.23&78.25&78.17&78.68&79.17&79.70&84.72&79.61&78.00&80.38&80.85&80.71\\
CapeFormer \cite{CapeFormer}&89.45&84.88&83.59&83.53&85.09 &85.31&91.94&88.92&89.40&88.01&88.25&89.30 \\ \cdashline{1-13}[1.5pt/3pt]
MetaPoint &83.45&79.88&77.59&76.53&79.09&79.31&84.13&80.30&78.02&77.12&79.58& 79.83\\
MetaPoint+ &\textbf{90.43}&\textbf{85.59}&\textbf{84.52}&\textbf{84.34}&\textbf{85.96}&\textbf{86.17}&\textbf{92.58}&\textbf{89.63}&\textbf{89.98}&\textbf{88.70}&\textbf{89.20}&\textbf{90.02} \\ \bottomrule
\end{tabular}
}
\end{table*}
\begin{table}  
\centering  
\caption{The evaluation results using mPCK metric.}  
\label{tab:mPCK}
\vspace{-0.25cm}
\begin{subtable}{0.5\textwidth}  
\centering  
\caption{in 1-shot and 5-shot settings.}  
\scalebox{0.85}{
\begin{tabular}{ccccccc}
\toprule 
                        &            & Split1 & Split2 & Split3 & Split4 & Split5 \\
\midrule                 
\multirow{2}{*}{1-shot} & CapeFormer &75.13&69.30&68.59&68.50&71.38 \\
                        & MetaPoint+ &\textbf{77.11}&\textbf{71.07}&\textbf{70.32}&\textbf{69.93}&\textbf{72.73} \\
\midrule                 
\multirow{2}{*}{5-shot} & CapeFormer &78.05&74.43&74.87&73.80&76.11   \\
                        & MetaPoint+ &\textbf{79.22}&\textbf{75.51}&\textbf{76.20}&\textbf{75.92}&\textbf{77.65}\\
\bottomrule 
\end{tabular}
}
\end{subtable}
~   
\begin{subtable}{0.5\textwidth}  
\centering  
\vspace{0.35cm}
\caption{in cross super-category pose estimation.}  
\scalebox{0.85}{
\begin{tabular}{ccccc} 
\toprule 
    & Human Body & Human Face & Vehicle & Furniture \\ 
\midrule  
CapeFormer &60.21&58.04&27.34&35.60 \\
MetaPoint+ &\textbf{62.42}&\textbf{60.72}&\textbf{30.78}&\textbf{38.52} \\  
\bottomrule 
\end{tabular}
}
\end{subtable}  
\end{table}  

\subsection{Comparison with Prior Works}
Following previous works \cite{CapeFormer}, our comparison baselines include ProtoNet \cite{ProtoNet}, MAML \cite{MAML}, Fine-tune \cite{Finetuning}, POMNet \cite{POMNet}, and CapeFormer \cite{CapeFormer}.
Specifically, ProtoNet, MAML, and Fine-tune are popular few-shot learning methods adapted in \cite{POMNet}.
POMNet \cite{POMNet} adopts a keypoint matching framework to match support keypoints to the pixels of query image.
CapeFormer \cite{CapeFormer} enhances above similarity modeling and further refines keypoints via transformer modules.
We use two versions of our model for comparison, \emph{i.e.}, MetaPoint and MetaPoint+.
MetaPoint directly employs the predicted meta-points as results, where the support keypoints are only used to determine the identities for evaluation. 
MetaPoint+ is the full-fledged version of our model and employs the refined meta-points as results.

We summarize the results of PCK@0.2 in 1-shot setting and 5-shot setting in Tab.~\ref{tab:SOTA}.
Firstly, we could see that our MetaPoint achieves satisfactory performances without any support and refinement, which indicates the mined meta-points could reveal the keypoints inherency and serve as high-quality proposals.
Our MetaPoint+ achieves the optimal performances against all baselines, demonstrating the effectiveness of our framework on learning and refining meta-points.
We could also see that our methods have relatively robust performances on various dataset splits and different number of support images.

In addition, the performances of various methods approach to saturation, probably because PCK@0.2 employs a coarse threshold.
We also report the results of mPCK (with the thresholds in [0.05, 0.1, 0.15, 0.2]) against the most competitive baseline (\emph{i.e.}, CapeFormer) for a more comprehensive evaluation in Tab.~\ref{tab:mPCK}.
From the perspective of mPCK, we can see that our MetaPoint+ outperforms dramatically against the most competitive baseline, which more clearly demonstrates the effectiveness of our novel framework based on meta-point learning and refining.

\subsection{Cross Super-Category Pose Estimation}
\begin{table}[t]
\centering
\caption{Results of cross super-category pose estimation.
We summarize the PCK@0.2 results and leave mPCK results in Tab.~\ref{tab:mPCK}.}
\label{tab:cross_super}
\scalebox{0.85}{
\begin{tabular}{ccccc} \toprule
Method & Human Body & Human Face & Vehicle & Furniture\\ \midrule
ProtoNet &37.61&57.80&28.35&42.64\\
MAML &51.93&25.72&17.68&20.09\\
Fine-tune &52.11&25.53&17.46&20.76\\
POMNet &73.82&79.63&34.92&47.27\\
CapeFormer & 83.44 & 80.96 & 45.40 & 52.49 \\ 
MetaPoint+ &\textbf{84.32}&\textbf{82.21}&\textbf{46.51}&\textbf{53.67} \\    \bottomrule
\end{tabular}
}
\end{table}

\begin{figure*}[ht]
\begin{center}
\includegraphics[width=0.93\linewidth]{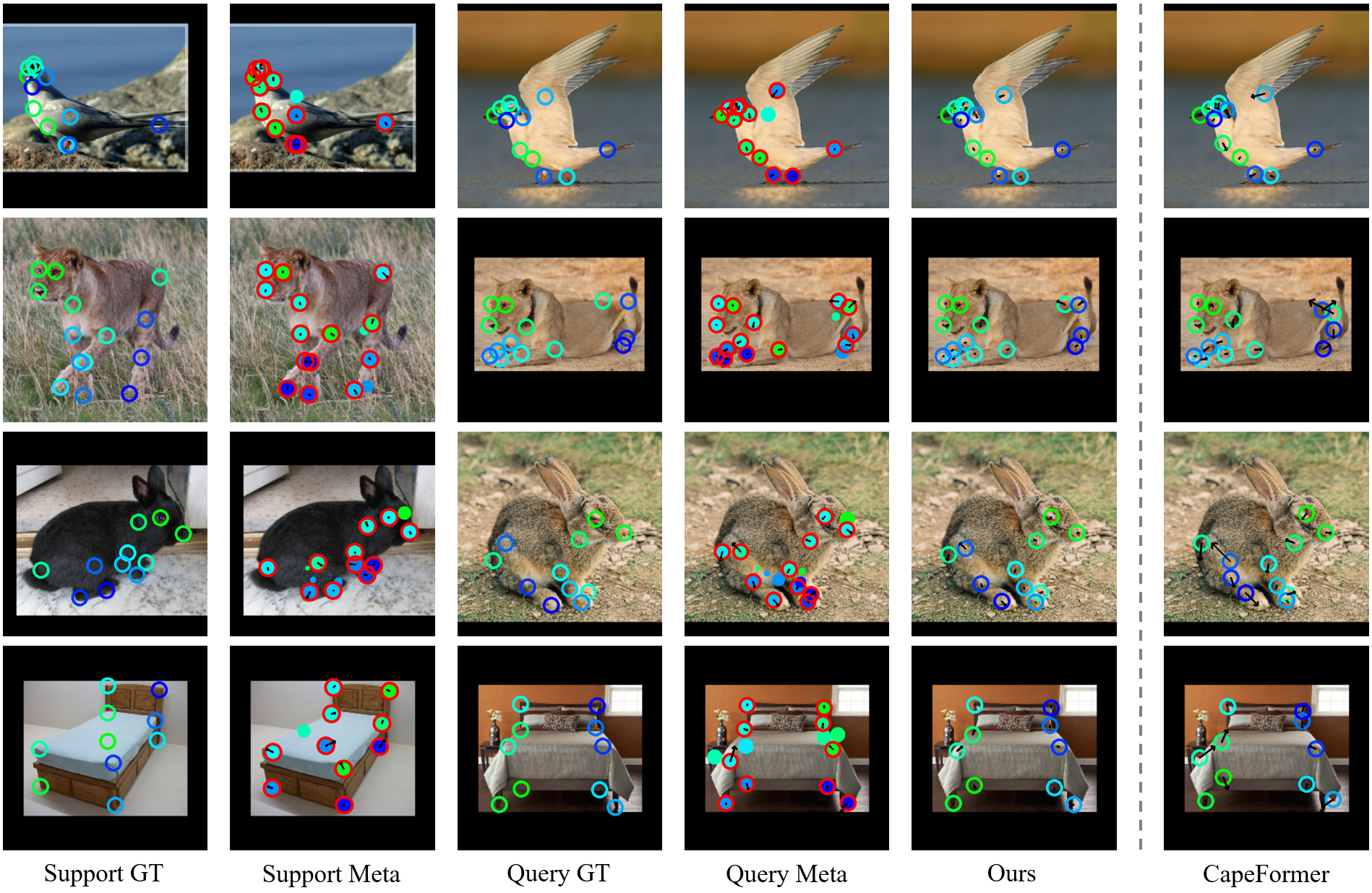}
\end{center}
\caption{
In each row, we show the GT keypoints and predictions of a pair of support image and query image. 
The left two columns show the GT keypoints and estimated meta-points on support image. 
The middle two columns show the GT keypoints and estimated meta-points on query image.
The right two columns show the final keypoints predicted by ours and CapeFormer.
We employ the small black arrows to indicate the deviations to GT.
The radii of drew meta-points indicate their visibilities, and the assigned meta-points are encircled with red.
}
\label{fig:viz_SOTA}
\end{figure*}

\begin{figure*}[ht]
\begin{center}
\includegraphics[width=0.94\linewidth]{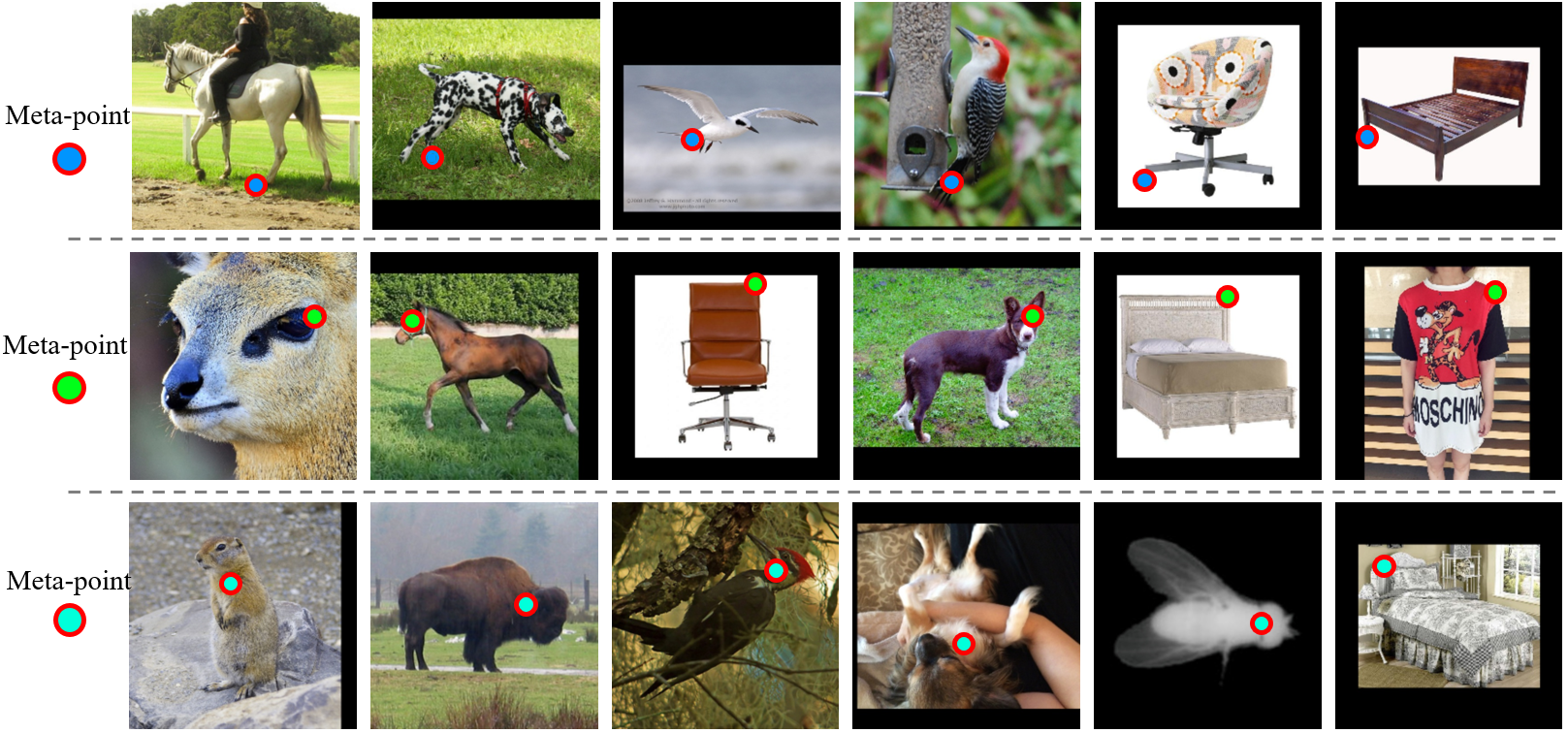}
\end{center}
\caption{
In each row, we track one meta-embedding and visualize its meta-point on various samples if the visibilities are greater than $0.5$.
}
\label{fig:viz_tracking}
\end{figure*}

In the standard setting of CAPE, although the training classes, validation classes, and test classes have no overlap, some classes may have common characteristics and thus easy to transfer, \emph{e.g.}, the body of various kinds of animals.
We follow the experiments of cross super-category evaluation in previous works \cite{CapeFormer,POMNet} to investigate the generalization ability on significantly different classes of our method.
Specifically, we respectively employ one of four super-categories (\emph{i.e.},  human face, human body, vehicle, and furniture) in MP-100 dataset as the test categories, while the other categories are used as the training categories.
As shown in Tab.~\ref{tab:cross_super}, our method outperforms than all baselines, demonstrating that the our method could mine more class-agnostic information via learning meta-points, and thus are more robust against the distribution shift of categories.
We also include the evaluation results using mPCK metric in Tab.~\ref{tab:mPCK}.
The performances vary greatly across super-classes, probably because different super-classes have diverse kinds of keypoints as discussed in \cite{CapeFormer}.

\subsection{Qualitative Analyses}

For a more intuitive understanding of our method, we visualize the predicted meta-points and keypoints in 1-shot setting for novel classes in Fig.~\ref{fig:viz_SOTA}.
As shown in the second column and the fourth column, our model could predict meaningful meta-points without any support, which have satisfactory performances compared with assigned GT.
Furthermore, the meta-points correspond to consistent semantic parts in support images and query images.
Therefore, the predicted meta-points not only reveal the inherency of keypoints but also could serve as favourable proposals for subsequent refinement.
As shown in the first row, the support keypoints of bird feet are hard to provide clear support information.
In such case, previous works may be confused by the ambiguous support, while our method could predict precise keypoints based on the inherent and complementary information mined via meta-points.
Similar cases could be found in other visualizations.

To further explore the properties of learned meta-points, we track one meta-embedding and visualize its meta-point on various samples if the visibilities are greater than $0.5$ in Fig.~\ref{fig:viz_tracking}.
As shown in the first row, we could find that the meta-point could correspond to consistent semantic parts of various classes, even if the parts have just high-level or potential relation.
Note that there is no annotation binding the points across classes, which demonstrates that our model could mine and extract the inherent and universal information of keypoints.
In the second row, the meta-point may correspond to the egoistic left-upper point of objects.
Similar phenomena could be found in other cases.

\subsection{Ablation Study}
\begin{table}[t]
\centering
\caption{Results (mPCK) of various component combinations.}
\label{tab:ablation}
\scalebox{1}{
\begin{tabular}{c|cccc|c}
\Xhline{0.8pt}
&  $+\Tilde{\bm{v}}$ & $+ \bm{F}_s$ & $+ \delta(\Tilde{\bm{E}_s})$ &  $+\mathcal{L}_{sl_1}$ & Split1 \\ \Xhline{0.4pt}
\#1 & - & - & - & -                                             & 68.18 \\ 
\#2 & \checkmark & - & - & -                                    & 70.37 \\
\#3 &\checkmark & \checkmark & - & -                            & 74.70\\
\#4 &\checkmark & \checkmark & \checkmark & -                   & 76.43\\
\#5 &\checkmark & \checkmark & \checkmark & \checkmark          & 77.11 \\ \Xhline{0.8pt}
\end{tabular}
}
\end{table}
\begin{table}[t]
\centering
\caption{Results (mPCK) of various configurations of our model.}
\label{tab:cfg}
\scalebox{1}{
\begin{tabular}{cccccc} \Xhline{0.8pt}
     & Version & Split1 & Split2 & Split3 & Split4 \\ \Xhline{0.4pt}
\#1&Standard     &77.11&71.07&70.32&69.93\\  
\#2&Dual-Asm     &73.21&66.90&65.78&65.61\\
\#3&$R=2$        &76.73&70.66&70.02&69.64\\
\#4&$R=4$        &77.18&71.12&70.40&69.97\\
\#5&$L=2$        &76.30&70.22&69.38&69.21\\
\#6&$L=4$        &77.20&71.15&70.44&69.98\\  \Xhline{0.8pt}
\end{tabular}
}
\end{table}

In this section, we investigate the performances of component combinations and various configurations of our method.
Firstly, we gradually enable various components in our model and report the mPCKs in Tab.~\ref{tab:ablation}.
The row \#1 is a plain version to directly predict and refine meta-points without extra information, which shows the basic performance of learned meta-points.
In row \#2, we further enable visibility $\Tilde{\bm{v}}$, which avoids the wrong but accidental assignment to non-existent meta-points (\emph{e.g.}, occluded) and thus prompts the performance.
In row \#3 and row \#4, we enable support features $\bm{F}_s$ and support meta-embeddings $\delta(\Tilde{\bm{E}_s})$ in refinement, which provide critical and complementary information for refining meta-points to desired keypoints.
Row \#5 indicating the contribution of our slacked regression loss $\mathcal{L}_{sl_1}$, which could slack the supervision on immature predictions.
In addition, we explore the impact of various configurations in Tab.~\ref{tab:cfg}.
Specifically, Row \#1 indicates the standard performance of our model.
Row \#2 solves the assignment $\delta(\cdot)$ based on the sum of cost matrices of query and support images but obtain worse results, probably because only the cost matrix of support image is available in inference.
Row \#3 and row \#4 explore the impact of feature map number $R$, and our default configuration (\emph{i.e.}, $R=3$) could mine effective features from pyramid feature maps via our point decoder, but too shallow feature map only improves negligibly (\emph{i.e.}, $R=4$) probably because the features are too low-level.
The last two rows investigate the impact of decoder layer number, and our point decoder could progressively refine keypoints via multiple decoder layers.
The performances are gradually saturated, and our default choice (\emph{i.e.}, $L=3$) is appropriate and enough.
\section{Conclusion}
In this paper, we have proposed a novel framework for CAPE based on meta-point learning and refining.
Such meta-points not only reveal the inherency of keypoints but also serve as meaningful proposals for downstream tasks.
We also have proposed a progressive deformable point decoder and a slacked regression loss for better prediction and supervision.
The effectiveness of our framework has been demonstrated by in-depth experiments on MP-100 dataset.

\section*{Acknowledgements}
This work was supported in part by the National Natural
Science Foundation of China under Grants 62132006, in part by the Natural Science Foundation of Jiangxi Province of China under Grants 20223AEI91002, 20232BAB202001, and 20224BAB212012, and in part by the Project funded by China Posdoctoral Science Foundation under Grant 2022M721417.


\begin{thebibliography}{57}
\providecommand{\natexlab}[1]{#1}
\providecommand{\url}[1]{\texttt{#1}}
\expandafter\ifx\csname urlstyle\endcsname\relax
  \providecommand{\doi}[1]{doi: #1}\else
  \providecommand{\doi}{doi: \begingroup \urlstyle{rm}\Url}\fi

\bibitem[Alexe et~al.(2012)Alexe, Deselaers, and Ferrari]{PropBox1}
Bogdan Alexe, Thomas Deselaers, and Vittorio Ferrari.
\newblock Measuring the objectness of image windows.
\newblock \emph{IEEE Transactions on Pattern Analysis and Machine Intelligence}, 34\penalty0 (11):\penalty0 2189--2202, 2012.

\bibitem[Andriluka et~al.(2014)Andriluka, Pishchulin, Gehler, and Schiele]{humanpose1}
Mykhaylo Andriluka, Leonid Pishchulin, Peter Gehler, and Bernt Schiele.
\newblock 2d human pose estimation: New benchmark and state of the art analysis.
\newblock In \emph{CVPR}, pages 3686--3693, 2014.

\bibitem[Browatzki and Wallraven(2020)]{face1}
Bjorn Browatzki and Christian Wallraven.
\newblock 3fabrec: Fast few-shot face alignment by reconstruction.
\newblock In \emph{CVPR}, pages 6110--6120, 2020.

\bibitem[Cao et~al.(2019)Cao, Tang, Fang, Shen, Lu, and Tai]{animalpose1}
Jinkun Cao, Hongyang Tang, Hao-Shu Fang, Xiaoyong Shen, Cewu Lu, and Yu-Wing Tai.
\newblock Cross-domain adaptation for animal pose estimation.
\newblock In \emph{ICCV}, pages 9498--9507, 2019.

\bibitem[Carion et~al.(2020)Carion, Massa, Synnaeve, Usunier, Kirillov, and Zagoruyko]{DETR}
Nicolas Carion, Francisco Massa, Gabriel Synnaeve, Nicolas Usunier, Alexander Kirillov, and Sergey Zagoruyko.
\newblock End-to-end object detection with transformers.
\newblock In \emph{ECCV}, pages 213--229, 2020.

\bibitem[Chen et~al.(2021{\natexlab{a}})Chen, Niu, Liu, and Zhang]{m2}
Junjie Chen, Li Niu, Liu Liu, and Liqing Zhang.
\newblock Weak-shot fine-grained classification via similarity transfer.
\newblock In \emph{NeurIPS}, pages 7306--7318, 2021{\natexlab{a}}.

\bibitem[Chen et~al.(2021{\natexlab{b}})Chen, Niu, and Zhang]{m3}
Junjie Chen, Li Niu, and Liqing Zhang.
\newblock Depth privileged scene recognition via dual attention hallucination.
\newblock \emph{IEEE Transactions on Image Processing}, 30:\penalty0 9164--9178, 2021{\natexlab{b}}.

\bibitem[Chen et~al.(2022{\natexlab{a}})Chen, Niu, Zhou, Si, Qian, and Zhang]{m1}
Junjie Chen, Li Niu, Siyuan Zhou, Jianlou Si, Chen Qian, and Liqing Zhang.
\newblock Weak-shot semantic segmentation via dual similarity transfer.
\newblock In \emph{NeurIPS}, pages 32525--32536, 2022{\natexlab{a}}.

\bibitem[Chen et~al.(2023)Chen, Niu, Zhang, Si, Qian, and Zhang]{m4}
Junjie Chen, Li Niu, Jianfu Zhang, Jianlou Si, Chen Qian, and Liqing Zhang.
\newblock Amodal instance segmentation via prior-guided expansion.
\newblock In \emph{AAAI}, pages 313--321, 2023.

\bibitem[Chen et~al.(2022{\natexlab{b}})Chen, Ma, Liu, Chen, Cui, Wei, and Wang]{x1}
Runnan Chen, Yuexin Ma, Lingjie Liu, Nenglun Chen, Zhiming Cui, Guodong Wei, and Wenping Wang.
\newblock Semi-supervised anatomical landmark detection via shape-regulated self-training.
\newblock \emph{Neurocomputing}, 471:\penalty0 335--345, 2022{\natexlab{b}}.

\bibitem[Chen et~al.(2018)Chen, Wang, Peng, Zhang, Yu, and Sun]{heatmap2}
Yilun Chen, Zhicheng Wang, Yuxiang Peng, Zhiqiang Zhang, Gang Yu, and Jian Sun.
\newblock Cascaded pyramid network for multi-person pose estimation.
\newblock In \emph{CVPR}, pages 7103--7112, 2018.

\bibitem[Chen et~al.(2021{\natexlab{c}})Chen, Liu, Xu, Darrell, and Wang]{fsl-cls1}
Yinbo Chen, Zhuang Liu, Huijuan Xu, Trevor Darrell, and Xiaolong Wang.
\newblock Meta-baseline: Exploring simple meta-learning for few-shot learning.
\newblock In \emph{ICCV}, pages 9062--9071, 2021{\natexlab{c}}.

\bibitem[Cheng et~al.(2020{\natexlab{a}})Cheng, Xiao, Wang, Shi, Huang, and Zhang]{heatmap3}
Bowen Cheng, Bin Xiao, Jingdong Wang, Honghui Shi, Thomas~S Huang, and Lei Zhang.
\newblock Higherhrnet: Scale-aware representation learning for bottom-up human pose estimation.
\newblock In \emph{CVPR}, pages 5386--5395, 2020{\natexlab{a}}.

\bibitem[Cheng et~al.(2021)Cheng, Schwing, and Kirillov]{MaskFormer}
Bowen Cheng, Alex Schwing, and Alexander Kirillov.
\newblock Per-pixel classification is not all you need for semantic segmentation.
\newblock pages 17864--17875, 2021.

\bibitem[Cheng et~al.(2020{\natexlab{b}})Cheng, Chung, Tai, and Tang]{PropMaskD2}
Ho~Kei Cheng, Jihoon Chung, Yu-Wing Tai, and Chi-Keung Tang.
\newblock Cascadepsp: Toward class-agnostic and very high-resolution segmentation via global and local refinement.
\newblock In \emph{CVPR}, pages 8890--8899, 2020{\natexlab{b}}.

\bibitem[Fan et~al.(2020)Fan, Zhuo, Tang, and Tai]{fsl-det1}
Qi Fan, Wei Zhuo, Chi-Keung Tang, and Yu-Wing Tai.
\newblock Few-shot object detection with attention-rpn and multi-relation detector.
\newblock In \emph{CVPR}, pages 4013--4022, 2020.

\bibitem[Finn et~al.(2017)Finn, Abbeel, and Levine]{MAML}
Chelsea Finn, Pieter Abbeel, and Sergey Levine.
\newblock Model-agnostic meta-learning for fast adaptation of deep networks.
\newblock In \emph{ICML}, pages 1126--1135, 2017.

\bibitem[Ge et~al.(2021)Ge, Zhang, and Luo]{Metacloth}
Yuying Ge, Ruimao Zhang, and Ping Luo.
\newblock Metacloth: Learning unseen tasks of dense fashion landmark detection from a few samples.
\newblock \emph{IEEE Transactions on Image Processing}, 31:\penalty0 1120--1133, 2021.

\bibitem[Geng et~al.(2021)Geng, Sun, Xiao, Zhang, and Wang]{reg_DEKR}
Zigang Geng, Ke Sun, Bin Xiao, Zhaoxiang Zhang, and Jingdong Wang.
\newblock Bottom-up human pose estimation via disentangled keypoint regression.
\newblock In \emph{CVPR}, pages 14676--14686, 2021.

\bibitem[He et~al.(2016)He, Zhang, Ren, and Sun]{ResNet}
Kaiming He, Xiangyu Zhang, Shaoqing Ren, and Jian Sun.
\newblock Deep residual learning for image recognition.
\newblock In \emph{CVPR}, pages 770--778, 2016.

\bibitem[He et~al.(2023)He, Bharaj, Ferman, Rhodin, and Garrido]{FewShot3DKP}
Xingzhe He, Gaurav Bharaj, David Ferman, Helge Rhodin, and Pablo Garrido.
\newblock Few-shot geometry-aware keypoint localization.
\newblock In \emph{CVPR}, pages 21337--21348, 2023.

\bibitem[Hosang et~al.(2015)Hosang, Benenson, Doll{\'a}r, and Schiele]{PropBox3}
Jan Hosang, Rodrigo Benenson, Piotr Doll{\'a}r, and Bernt Schiele.
\newblock What makes for effective detection proposals?
\newblock \emph{IEEE Transactions on Pattern Analysis and Machine Intelligence}, 38\penalty0 (4):\penalty0 814--830, 2015.

\bibitem[Humayun et~al.(2014)Humayun, Li, and Rehg]{PropMask1}
Ahmad Humayun, Fuxin Li, and James~M Rehg.
\newblock Rigor: Reusing inference in graph cuts for generating object regions.
\newblock In \emph{CVPR}, pages 336--343, 2014.

\bibitem[Jaiswal et~al.(2021)Jaiswal, Wu, Natarajan, and Natarajan]{PropBoxD1}
Ayush Jaiswal, Yue Wu, Pradeep Natarajan, and Premkumar Natarajan.
\newblock Class-agnostic object detection.
\newblock In \emph{WCAV}, pages 919--928, 2021.

\bibitem[Kr{\"a}henb{\"u}hl and Koltun(2014)]{PropMask2}
Philipp Kr{\"a}henb{\"u}hl and Vladlen Koltun.
\newblock Geodesic object proposals.
\newblock In \emph{ECCV}, pages 725--739. Springer, 2014.

\bibitem[Labuguen et~al.(2021)Labuguen, Matsumoto, Negrete, Nishimaru, Nishijo, Takada, Go, Inoue, and Shibata]{animalpose2}
Rollyn Labuguen, Jumpei Matsumoto, Salvador~Blanco Negrete, Hiroshi Nishimaru, Hisao Nishijo, Masahiko Takada, Yasuhiro Go, Ken-ichi Inoue, and Tomohiro Shibata.
\newblock Macaquepose: a novel “in the wild” macaque monkey pose dataset for markerless motion capture.
\newblock \emph{Frontiers in behavioral neuroscience}, 14:\penalty0 581154, 2021.

\bibitem[Li et~al.(2019)Li, Wang, Zhu, Mao, Fang, and Lu]{humanpose2}
Jiefeng Li, Can Wang, Hao Zhu, Yihuan Mao, Hao-Shu Fang, and Cewu Lu.
\newblock Crowdpose: Efficient crowded scenes pose estimation and a new benchmark.
\newblock In \emph{CVPR}, pages 10863--10872, 2019.

\bibitem[Li et~al.(2021)Li, Bian, Zeng, Wang, Pang, Liu, and Lu]{reg_RLE}
Jiefeng Li, Siyuan Bian, Ailing Zeng, Can Wang, Bo Pang, Wentao Liu, and Cewu Lu.
\newblock Human pose regression with residual log-likelihood estimation.
\newblock In \emph{ICCV}, pages 11025--11034, 2021.

\bibitem[Liu et~al.(2020)Liu, Zhang, Lin, and Liu]{fsl-seg2}
Weide Liu, Chi Zhang, Guosheng Lin, and Fayao Liu.
\newblock Crnet: Cross-reference networks for few-shot segmentation.
\newblock In \emph{CVPR}, pages 4165--4173, 2020.

\bibitem[Lu and Koniusz(2022)]{FSKD}
Changsheng Lu and Piotr Koniusz.
\newblock Few-shot keypoint detection with uncertainty learning for unseen species.
\newblock In \emph{CVPR}, pages 19416--19426, 2022.

\bibitem[Lu et~al.(2023)Lu, Zhu, and Koniusz]{SFSKD}
Changsheng Lu, Hao Zhu, and Piotr Koniusz.
\newblock From saliency to dino: Saliency-guided vision transformer for few-shot keypoint detection.
\newblock \emph{arXiv preprint arXiv:2304.03140}, 2023.

\bibitem[Luo et~al.(2016)Luo, Li, Urtasun, and Zemel]{nonlocal2}
Wenjie Luo, Yujia Li, Raquel Urtasun, and Richard Zemel.
\newblock Understanding the effective receptive field in deep convolutional neural networks.
\newblock 2016.

\bibitem[Luo et~al.(2021)Luo, Wang, Huang, Wang, Tan, and Zhou]{heatmap_SAHR}
Zhengxiong Luo, Zhicheng Wang, Yan Huang, Liang Wang, Tieniu Tan, and Erjin Zhou.
\newblock Rethinking the heatmap regression for bottom-up human pose estimation.
\newblock In \emph{CVPR}, pages 13264--13273, 2021.

\bibitem[Mao et~al.(2022)Mao, Ge, Shen, Tian, Wang, Wang, and den Hengel]{query1}
Weian Mao, Yongtao Ge, Chunhua Shen, Zhi Tian, Xinlong Wang, Zhibin Wang, and Anton~van den Hengel.
\newblock Poseur: Direct human pose regression with transformers.
\newblock In \emph{ECCV}, pages 72--88. Springer, 2022.

\bibitem[Nakamura and Harada(2019)]{Finetuning}
Akihiro Nakamura and Tatsuya Harada.
\newblock Revisiting fine-tuning for few-shot learning.
\newblock \emph{arXiv preprint arXiv:1910.00216}, 2019.

\bibitem[Nie et~al.(2019)Nie, Feng, Zhang, and Yan]{reg1}
Xuecheng Nie, Jiashi Feng, Jianfeng Zhang, and Shuicheng Yan.
\newblock Single-stage multi-person pose machines.
\newblock In \emph{ICCV}, pages 6951--6960, 2019.

\bibitem[O~Pinheiro et~al.(2015)O~Pinheiro, Collobert, and Doll{\'a}r]{PropMaskDeep}
Pedro~O O~Pinheiro, Ronan Collobert, and Piotr Doll{\'a}r.
\newblock Learning to segment object candidates.
\newblock 2015.

\bibitem[Reddy et~al.(2018)Reddy, Vo, and Narasimhan]{carpose1}
N~Dinesh Reddy, Minh Vo, and Srinivasa~G Narasimhan.
\newblock Carfusion: Combining point tracking and part detection for dynamic 3d reconstruction of vehicles.
\newblock In \emph{CVPR}, pages 1906--1915, 2018.

\bibitem[Ren et~al.(2015)Ren, He, Girshick, and Sun]{FasterRCNN}
Shaoqing Ren, Kaiming He, Ross Girshick, and Jian Sun.
\newblock Faster r-cnn: Towards real-time object detection with region proposal networks.
\newblock 2015.

\bibitem[Shi et~al.(2022)Shi, Wei, Li, Ren, and Tan]{query_PETR}
Dahu Shi, Xing Wei, Liangqi Li, Ye Ren, and Wenming Tan.
\newblock End-to-end multi-person pose estimation with transformers.
\newblock In \emph{CVPR}, pages 11069--11078, 2022.

\bibitem[Shi et~al.(2023)Shi, Huang, Ma, Hu, and Cao]{CapeFormer}
Min Shi, Zihao Huang, Xianzheng Ma, Xiaowei Hu, and Zhiguo Cao.
\newblock Matching is not enough: A two-stage framework for category-agnostic pose estimation.
\newblock In \emph{CVPR}, pages 7308--7317, 2023.

\bibitem[Snell et~al.(2017)Snell, Swersky, and Zemel]{ProtoNet}
Jake Snell, Kevin Swersky, and Richard Zemel.
\newblock Prototypical networks for few-shot learning.
\newblock 2017.

\bibitem[Song et~al.(2019)Song, Wang, Zhou, Zhu, Guan, Dai, Su, Li, and Yang]{carpose2}
Xibin Song, Peng Wang, Dingfu Zhou, Rui Zhu, Chenye Guan, Yuchao Dai, Hao Su, Hongdong Li, and Ruigang Yang.
\newblock Apollocar3d: A large 3d car instance understanding benchmark for autonomous driving.
\newblock In \emph{CVPR}, pages 5452--5462, 2019.

\bibitem[Vaswani et~al.(2017)Vaswani, Shazeer, Parmar, Uszkoreit, Jones, Gomez, Kaiser, and Polosukhin]{multiheadattention}
Ashish Vaswani, Noam Shazeer, Niki Parmar, Jakob Uszkoreit, Llion Jones, Aidan~N Gomez, {\L}ukasz Kaiser, and Illia Polosukhin.
\newblock Attention is all you need.
\newblock 2017.

\bibitem[Wang et~al.(2018)Wang, Girshick, Gupta, and He]{nonlocal}
Xiaolong Wang, Ross Girshick, Abhinav Gupta, and Kaiming He.
\newblock Non-local neural networks.
\newblock In \emph{CVPR}, pages 7794--7803, 2018.

\bibitem[Wang et~al.(2020)Wang, Yao, Kwok, and Ni]{fsl1}
Yaqing Wang, Quanming Yao, James~T Kwok, and Lionel~M Ni.
\newblock Generalizing from a few examples: A survey on few-shot learning.
\newblock \emph{ACM computing surveys (csur)}, 53\penalty0 (3):\penalty0 1--34, 2020.

\bibitem[Wei et~al.(2021)Wei, Liu, Wang, and Tai]{face2}
Zhen Wei, Bingkun Liu, Weinong Wang, and Yu-Wing Tai.
\newblock Few-shot model adaptation for customized facial landmark detection, segmentation, stylization and shadow removal.
\newblock \emph{arXiv preprint arXiv:2104.09457}, 2021.

\bibitem[Xu et~al.(2022{\natexlab{a}})Xu, Jin, Zeng, Liu, Qian, Ouyang, Luo, and Wang]{POMNet}
Lumin Xu, Sheng Jin, Wang Zeng, Wentao Liu, Chen Qian, Wanli Ouyang, Ping Luo, and Xiaogang Wang.
\newblock Pose for everything: Towards category-agnostic pose estimation.
\newblock In \emph{ECCV}, pages 398--416, 2022{\natexlab{a}}.

\bibitem[Xu et~al.(2022{\natexlab{b}})Xu, Zhang, Wei, Lin, Cao, Hu, and Bai]{PropMaskD1}
Mengde Xu, Zheng Zhang, Fangyun Wei, Yutong Lin, Yue Cao, Han Hu, and Xiang Bai.
\newblock A simple baseline for open-vocabulary semantic segmentation with pre-trained vision-language model.
\newblock In \emph{ECCV}, pages 736--753. Springer, 2022{\natexlab{b}}.

\bibitem[Xu et~al.(2022{\natexlab{c}})Xu, Zhang, Zhang, and Tao]{query2}
Yufei Xu, Jing Zhang, Qiming Zhang, and Dacheng Tao.
\newblock Vitpose: Simple vision transformer baselines for human pose estimation.
\newblock pages 38571--38584, 2022{\natexlab{c}}.

\bibitem[Yang and Ramanan(2012)]{PCK}
Yi Yang and Deva Ramanan.
\newblock Articulated human detection with flexible mixtures of parts.
\newblock \emph{IEEE Transactions on Pattern Analysis and Machine Intelligence}, 35\penalty0 (12):\penalty0 2878--2890, 2012.

\bibitem[Yin et~al.(2022)Yin, Gong, Wang, Yu, and Wang]{x0}
Zihao Yin, Ping Gong, Chunyu Wang, Yizhou Yu, and Yizhou Wang.
\newblock One-shot medical landmark localization by edge-guided transform and noisy landmark refinement.
\newblock In \emph{ECCV}, pages 473--489. Springer, 2022.

\bibitem[Zhang et~al.(2019)Zhang, Lin, Liu, Yao, and Shen]{fsl-seg1}
Chi Zhang, Guosheng Lin, Fayao Liu, Rui Yao, and Chunhua Shen.
\newblock Canet: Class-agnostic segmentation networks with iterative refinement and attentive few-shot learning.
\newblock In \emph{CVPR}, pages 5217--5226, 2019.

\bibitem[Zhang et~al.(2020)Zhang, Zhu, Dai, Ye, and Zhu]{heatmap1}
Feng Zhang, Xiatian Zhu, Hanbin Dai, Mao Ye, and Ce Zhu.
\newblock Distribution-aware coordinate representation for human pose estimation.
\newblock In \emph{CVPR}, pages 7093--7102, 2020.

\bibitem[Zhang and Wang(2021)]{fsl-det2}
Weilin Zhang and Yu-Xiong Wang.
\newblock Hallucination improves few-shot object detection.
\newblock In \emph{CVPR}, pages 13008--13017, 2021.

\bibitem[Zhu et~al.(2021)Zhu, Su, Lu, Li, Wang, and Dai]{DeformDETR}
Xizhou Zhu, Weijie Su, Lewei Lu, Bin Li, Xiaogang Wang, and Jifeng Dai.
\newblock Deformable detr: Deformable transformers for end-to-end object detection.
\newblock In \emph{ICLR}, 2021.

\bibitem[Zitnick and Doll{\'a}r(2014)]{PropBox2}
C~Lawrence Zitnick and Piotr Doll{\'a}r.
\newblock Edge boxes: Locating object proposals from edges.
\newblock In \emph{ECCV}, pages 391--405. Springer, 2014.

\end{thebibliography}

{\small
\bibliographystyle{ieeenat_fullname}

}

\end{document}